\documentclass{article}

\usepackage[dblblindworkshop, final, nonatbib]{neurips_2026}

\usepackage[
    backend=biber,
    style=numeric-comp,
    sorting=none,
    maxcitenames=1,
    maxbibnames=2,
]{biblatex}

\usepackage{amsmath,amsfonts,bm}

\def\eqref#1{equation~\ref{#1}}

\def\1{\bm{1}}

\def\vmu{{\bm{\mu}}}

\def\vf{{\bm{f}}}
\def\vg{{\bm{g}}}
\def\vh{{\bm{h}}}

\def\vu{{\bm{u}}}

\def\vx{{\bm{x}}}

\def\vz{{\bm{z}}}
\def\vF{{\bm{F}}}
\def\vG{{\bm{G}}}

\def\vsigma{{\bm{\sigma}}}
\def\vepsilon{{\bm{\varepsilon}}}

\DeclareMathAlphabet{\mathsfit}{\encodingdefault}{\sfdefault}{m}{sl}
\SetMathAlphabet{\mathsfit}{bold}{\encodingdefault}{\sfdefault}{bx}{n}

\newcommand{\diff}{\mathrm d}
\newcommand{\mat}[1]{\mathbf{#1}}

\usepackage{enumitem}  
\usepackage[utf8]{inputenc} %
\usepackage[T1]{fontenc}    %
\usepackage{hyperref}       %
\usepackage{url}            %
\usepackage{cleveref}
\usepackage{subcaption}
\usepackage{booktabs}       %
\usepackage{amsfonts}       %
\usepackage{nicefrac}       %
\usepackage{microtype}      %
\usepackage{xcolor}         %
\usepackage{tabularx}
\usepackage{adjustbox}
\usepackage{multirow}
\usepackage{float}
\usepackage{algorithm} 
\usepackage{algorithmic}

\newcommand{\mycomment}[1]{\textbf{\small\color{red}{\# #1}}}
\newcommand{\hl}[2]{\colorbox{#1!20}{$\displaystyle #2$}}

\workshoptitle{AI for Stochastic Dynamics / Sim2Science}

\title{SDECast: Probabilistic Weather Forecasting in Continuous Time with Neural SDEs}

\author{
Maria Marchenko\thanks{Equal contribution} \\
University of Amsterdam \\
\texttt{maria.m.m.marchenko@gmail.com} \\
\And
Martin Andrae$\footnotemark[1]$  \\
Linköping University, Sweden \\
\texttt{martin.andrae@liu.se} \\
\AND
Fredrik Lindsten\thanks{Equal advising} \\
Linköping University, Sweden \\
\texttt{fredrik.lindsten@liu.se} \\
\And
Christian A. Naesseth$\footnotemark[2]$ \\
University of Amsterdam \\
\texttt{christian.a.naesseth@uva.nl}
}

\begin{document}

\maketitle

\begin{abstract}
Existing machine learning weather forecasting models typically generate forecasts through autoregressive rollouts at a fixed temporal resolution.
While highly efficient for long-range prediction, this formulation can suffer from severe error accumulation when used with shorter time steps and does not explicitly encode the locality and temporal continuity of atmospheric dynamics.
To address these limitations, we introduce \textbf{SDECast}, a Neural Stochastic Differential Equation (SDE) framework for continuous-time probabilistic weather forecasting.
SDECast extends SDE Matching to learn stochastic dynamics directly in physical space, without requiring repeated SDE simulation during training.
On a simulated geophysical flow, we show that SDECast recovers meaningful drift dynamics and faithfully reproduces the underlying continuous-time behavior. 
We then demonstrate its scalability to global weather forecasting at hourly resolution, where SDECast produces skillful probabilistic forecasts for lead times of up to five days.

\end{abstract}

\section{Introduction}
Weather forecasting is a problem with broad societal impact: it informs early warning systems, supports energy planning, and shapes the daily choices of billions of people. While physics-based Numerical Weather Prediction (NWP) has been the dominant approach for decades, data-driven methods have recently matched or exceeded NWP accuracy at a fraction of the computational cost. This progress spans both deterministic and probabilistic forecasting, with state-of-the-art models relying on generative methods to model the distribution of possible atmospheric states \cite{bi2023accurate, lam2023learning, price2024gencast, alet2025fgn}. 

Despite this progress, current machine-learning approaches have several limitations. Unlike physics-based models, which evolve the atmosphere by solving Partial Differential Equations (PDEs), Machine Learning Weather Prediction (MLWP) models rely on autoregressive rollouts of a learned one-step predictor $p_\theta(\vx_t|\vx_{t-1})$. While this enables efficient long-range forecasting, it can suffer from severe error accumulation when applied with shorter timesteps \cite{AndraeLOL25}. Moreover, it fails to exploit the inductive biases of locality and temporal continuity that are naturally encoded in differential equations. Neural Ordinary Differential Equations (ODEs) have been explored as an alternative for weather modeling, but they are deterministic \cite{verma2024climode, seong2026rainode, he2025flowcast} or expensive to train due to their simulation-based loss \cite{Kochkov2024}. 

A related but relatively unexplored class of models is Neural Stochastic Differential Equations (SDEs). Like Neural ODEs, these models parameterize the system dynamics through a drift term, but additionally introduce stochasticity via a diffusion term, enabling principled probabilistic forecasting and efficient ensemble generation. 
Neural SDEs have traditionally been learned from either finely resolved trajectories \cite{craigmile2023statistical} or objectives that require repeated SDE simulation during training \cite{li2020scalable}, limiting their application to high-dimensional systems.
Recently, SDE Matching \cite{bartosh2025sde} introduced a simulation-free objective for learning latent SDEs from discrete time observations. 
In this work, we adapt this framework to learn stochastic dynamics directly in physical space and investigate its application to high-dimensional weather forecasting.

\subsection{Contributions}
We present \textbf{SDECast}, a Neural SDE framework for probabilistic weather forecasting. In summary, the strength and novelty of SDECast lie in its following capabilities:
\begin{enumerate}[label=(\roman*)]
\item \textbf{Continuous-time forecasting}, enabling ensemble generation at arbitrary lead times through stochastic trajectories from a single learned dynamical model.
\item \textbf{Interpretable physical-space dynamics}, encoding the locality and temporal continuity inherent to differential equations through explicit drift and diffusion functions.
\item \textbf{Simulation-free training}, scaling Neural SDE learning to high-dimensional weather forecasting without repeated SDE simulation during optimization.
\end{enumerate}

\section{Method}
SDECast adapts SDE Matching \cite{bartosh2025sde} to learn stochastic continuous-time dynamics in a scalable way. 
Our objective is to learn an SDE that governs the evolution of the physical state $\vx_t$. 
However, the simulation-free training objective of SDE Matching is formulated in terms of an auxiliary latent process $\vz_t$. 
We therefore introduce such a process to construct the training objective, but constrain it to coincide with the physical state at observation times. 
This makes the learned prior SDE directly interpretable as a model of the physical dynamics.

We first describe the simulation-free SDE Matching framework and then introduce our physical-space posterior parameterization.

\subsection{Simulation-free SDE Learning}
Let $\mat{X} = \{\vx_{t_i}\}_{i=0}^{N}$, $0 = t_0 < t_1 < \cdots < t_N = T$
denote a discretely observed trajectory of the physical system. We introduce a latent process $\vz_t$ governed by the {\em Prior SDE} 
\begin{align} 
\diff \vz_t = {\vh_\theta(\vz_t,t)}\,\diff t + {\vg_\theta(\vz_t,t)}\,\diff W_t, \label{eq:prior-sde} 
\end{align}
where $\vh_\theta$ and $\vg_\theta$ parameterize the drift and diffusion, respectively. 
This is the SDE used to generate forecasts at inference times. Following SDE Matching, we introduce the {\em Posterior SDE}
\begin{align} 
\diff \vz_t = \vf_{\varphi,\theta}(\vz_t,t;\mat{X})\,\diff t + \vg_\theta(\vz_t,t)\,\diff W_t,
\label{eq:posterior-sde} 
\end{align}
which shares its diffusion coefficient with the prior SDE, but conditions its drift on the observed trajectory $\mat{X}$. 
The posterior SDE can be interpreted as a variational approximation of the true posterior over trajectories induced by the prior \eqref{eq:prior-sde} coupled with an observation likelihood 
$p_\theta(\vx_{t}\mid\vz_{t})$.
It can also be seen as a controlled version of the prior SDE, modifying its dynamics to be consistent with the observed trajectory $\mat{X}$. 
Since the prior and posterior share the same diffusion coefficient, this control $\vu_{\theta,\varphi}$ is defined implicitly by
\begin{align} 
\vf_{\varphi,\theta}(\vz_t,t;\mat{X}) =  \vh_\theta(\vz_t,t) + \vg_\theta(\vz_t,t) \vu_{\theta,\varphi}(\vz_t,t,\mat{X}) . \label{eq:control} 
\end{align}
The variational objective penalizes the norm of the control while also reconstructing the observations:
\begin{align} 
\mathcal{L}_{\mathrm{dyn}} = \underset{q_\varphi(\mat{Z}|\mat{X})}{\mathbb{E}} \left[ \int_0^T \frac{1}{2} \left\| \vu_{\theta,\varphi}(\vz_t,t,\mat{X}) \right\|_2^2 \,\diff t \right], \quad
 \mathcal{L}_{\mathrm{rec}} =  \underset{q_\varphi(\mat{Z}|\mat{X})}{\mathbb{E}} \left[ \sum_{i=0}^{N} -\log p_\theta(\vx_{t_i}\mid\vz_{t_i}) \right]. 
 \label{eq:loss} 
\end{align}
Both terms are minimized jointly with respect to $\theta$ and $\varphi$. Direct evaluation of this objective would normally require repeatedly simulating the posterior SDE. SDE Matching avoids this by parameterizing its marginals $q_\varphi(\vz_t|\mat{X})$ directly as
\begin{align}
    \vz_t = \vF_\varphi(\vepsilon,t,\mat{X}), \qquad \vepsilon\sim\mathcal{N}(\mathbf{0},\mat{I}). \label{eq:posterior-map}
\end{align}
Assuming that $\vF_\varphi$ is differentiable with respect to $t$ and $\vepsilon$, and invertible with respect to $\vepsilon$, this induces an analytical expression for the posterior drift $\vf_{\varphi,\theta}$ in \eqref{eq:control}.
Consequently, the loss in \eqref{eq:loss} can be estimated by sampling $(t,\vepsilon)$ and evaluating $\vF_\varphi$, without integrating the posterior SDE. For more details on this we refer to Appendix~\cref{app:sde-matching}.

\subsection{Physical-space SDE}
The general SDE Matching formulation remains a latent-variable model: the latent state $\vz_t$ need not correspond to the physical state, and the posterior $q_\varphi(\vz_t\mid\mat{X})$ may depend on the complete observed trajectory. We instead seek an SDE whose state can be identified with the physical state of the system.

To establish this connection, we choose the highly concentrated observation model
\begin{align} 
p_\theta(\vx_t\mid\vz_t) = \mathcal{N}(\vx_t;\vz_t,\delta^2\mat{I}), \label{eq:observation-model} 
\end{align}
where $\delta\ll1$ is a fixed, small constant. For sufficiently small $\delta$, the posterior at an observation time concentrates near the corresponding physical state. Consequently, the posterior between two consecutive observations is expected to depend predominantly on the observations that bracket that interval, rather than on the entire trajectory. Motivated by this, we restrict the variational family $q_\varphi$, imposing a Markovian conditional independence structure:
\begin{align} 
q_\varphi(\vz_t\mid\mat{X}) = q_\varphi( \vz_t\mid \vx_{t_i},\vx_{t_{i+1}} ), \quad t\in(t_i,t_{i+1}), \qquad
q_\varphi(\vz_{t_i}\mid\mat{X}) = q_\varphi( \vz_{t_i}\mid \vx_{t_i}).
\label{eq:local-posterior} 
\end{align}
This restriction avoids conditioning on the complete trajectory and allows the posterior to be evaluated independently on each observation interval.

For $t\in[t_i,t_{i+1}]$, define the normalized time $\alpha_t = \frac{t-t_i}{t_{i+1}-t_i}$. 
We realize the local variational family using a Gaussian posterior
\begin{align} 
\vF_\varphi( \vepsilon,t, \mat{X}) 
&= (1-\alpha_t)\vx_{t_i} + \alpha_t\vx_{t_{i+1}} 
+ \delta\vepsilon \nonumber \\
&+ \alpha_t(1-\alpha_t) \left[ 
\vmu_\varphi(t,\vx_{t_i},\vx_{t_{i+1}}) 
+ \vsigma_\varphi(t,\vx_{t_i},\vx_{t_{i+1}}) \vepsilon
\right]
\label{eq:local-posterior-map} 
\end{align}
The linear term connects the neighboring observations, while $\vmu_\varphi$ and $\vsigma_\varphi$ model the conditional mean correction and uncertainty between them. The factor $\alpha_t(1-\alpha_t)$ makes both corrections vanish at the interval endpoints, ensuring that $\vz_{t_i}\sim\mathcal{N}(\vx_{t_i},\delta^2 \mat{I})$. This anchors the variational posterior to the concentrated observation model at each observation time, while allowing flexible stochastic dynamics between observations. Inserting this parametrization into the reconstruction term yields
\begin{align} 
\mathcal{L}_{\mathrm{rec}} &= \sum_{i=0}^{N} \mathbb{E}_{q_\varphi(\vz_{t_i}\mid\mat{X})} \left[ \frac{1}{2\delta^2} \left\| \vx_{t_i}-\vz_{t_i} \right\|_2^2 + \frac{d}{2}\log(2\pi\delta^2) \right] = \frac{(N+1)d}{2}\left(1+\log(2\pi\delta^2)\right).
\end{align}
Since $\delta$ is fixed, this term is constant with respect to the learnable parameters and can therefore be omitted during optimization. 
At inference time, we initialize the prior SDE in \eqref{eq:prior-sde} from the latest observed physical state and simulate it forward. Its trajectories provide forecasts at arbitrary lead times, while independent Wiener paths produce probabilistic forecast ensembles.

\section{Experiments}
To assess its performance across both simulated and real-world settings, we evaluate SDECast on the Surface Quasi-Geostrophic (SQG) system \cite{tulloch2009qgturbulence} and on global weather forecasting using ERA5 \cite{hersbach2020era5}. Further experimental and implementation details are provided in Appendix~\cref{app:experiments}.

\subsection{Surface Quasi-Geostrophic Dynamics}
We first evaluate SDECast on the SQG system, a widely used model of turbulent geophysical flows. Despite its relative simplicity, SQG captures key properties of geophysical turbulence, including sensitivity to initial conditions and interactions across multiple scales, while remaining computationally tractable \cite{tulloch2009qgturbulence}. 
Moreover, because the data are generated by a known numerical simulator, this setting allows us to evaluate not only forecast quality but also whether SDECast recovers the underlying continuous-time dynamics.

We compare SDECast with an autoregressive conditional flow-matching model (CFM). We also consider an ablation in which the learnable corrections $\vmu_\varphi$ and $\vsigma_\varphi$ are removed from $\vF_\varphi$, such that the prior SDE is learned against a fixed linear variational posterior. Since the SQG system is autonomous, we remove the explicit time conditioning from the prior drift and diffusion. We parameterize $\vg_\theta$ and $\vsigma_\varphi$ as diagonal matrices, and further restrict $\vg_\theta$ to be constant and independent of the state. 

\begin{figure}[]
  \centering
  \setlength{\baselineskip}{0pt}
  \setlength{\lineskip}{0pt}
\includegraphics[width=\textwidth, trim={0 0 0 0}, clip]{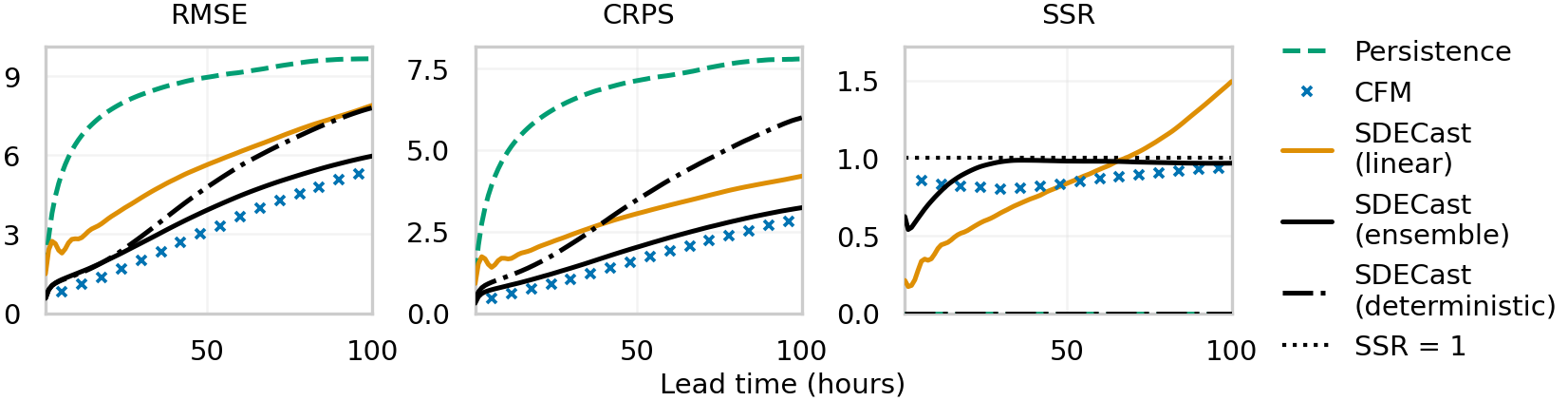}
  \caption{The RMSE ($\downarrow$), CRPS ($\downarrow$), SSR for experiments on SQG.}
  \label{fig:sqg-rmse-crps-ssr}
\end{figure}

We train all models using snapshots separated by six hours, but evaluate their forecasts at one-hour resolution. As shown in \cref{fig:sqg-rmse-crps-ssr}, SDECast produces stable and skillful rollouts over multiple forecast steps. Compared with the autoregressive flow-matching baseline, SDECast obtains higher RMSE and CRPS, but maintains an almost ideal spread-skill ratio (SSR). Crucially, it generates forecasts at a substantially finer temporal resolution than that observed during training, whereas CFM is restricted to its trained resolution.
Replacing the learned posterior with the fixed linear posterior substantially degrades performance, highlighting the importance of a learnable posterior.

\begin{figure}[]
  \centering
\includegraphics[width=\textwidth, trim={0 0 0 0}, clip]{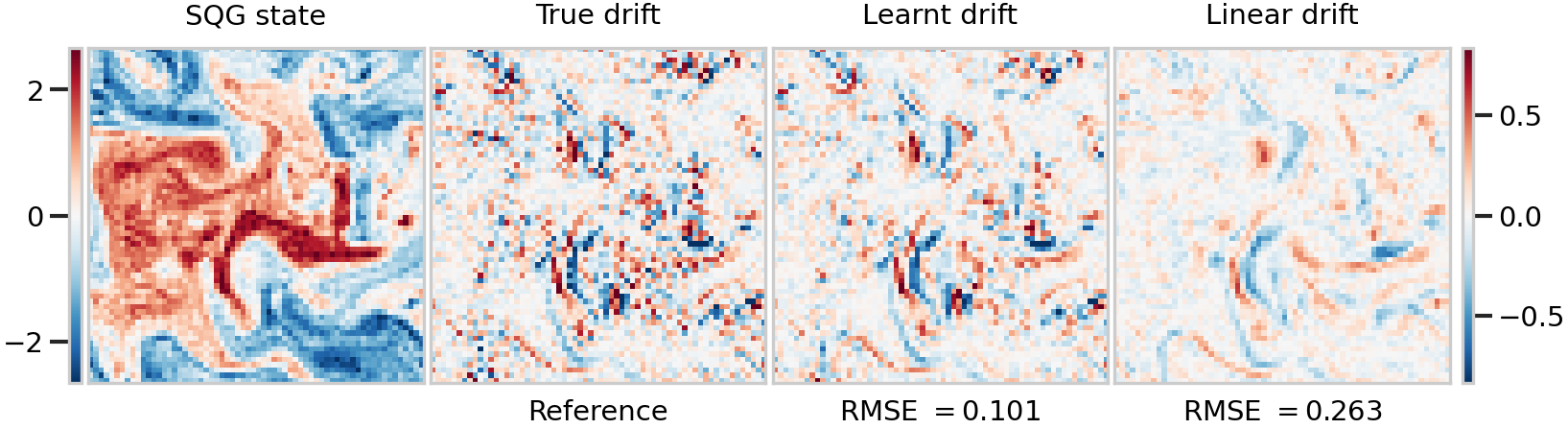}
  \caption{
  Comparison of the true drift, SDECast drift, and the six-hour finite-difference estimate. 
 }
  \label{fig:drift-sqg}
\end{figure}

Access to the numerical SQG simulator allows us to compare the learned prior drift with the true drift. 
In \cref{fig:drift-sqg}, we compare both quantities alongside a six-hour finite-difference estimate. 
The learned drift closely resembles the true drift and provides a more faithful description of the instantaneous dynamics than the finite-difference estimate, indicating that SDECast can recover continuous-time behavior from coarsely sampled observations. 
The learned drift nevertheless underrepresents the highest spatial frequencies, as expected: under coarse temporal sampling the fastest dynamics are only weakly resolved, and SDECast may attribute part of this high-frequency variability to stochasticity rather than to the deterministic drift.
Thus, although the underlying SQG simulator is deterministic, the learned SDE provides an effective stochastic representation of its unresolved chaotic dynamics. 
Consistent with this, removing the diffusion at inference time, denoted SDECast (deterministic), substantially degrades forecast performance and produces overly smooth, mean-like trajectories.

\subsection{Global Weather Forecasting with ERA5}

\begin{figure}[]
  \centering
  \setlength{\baselineskip}{0pt}
  \setlength{\lineskip}{0pt}
\includegraphics[width=\textwidth, trim={0 0 0 0}, clip]{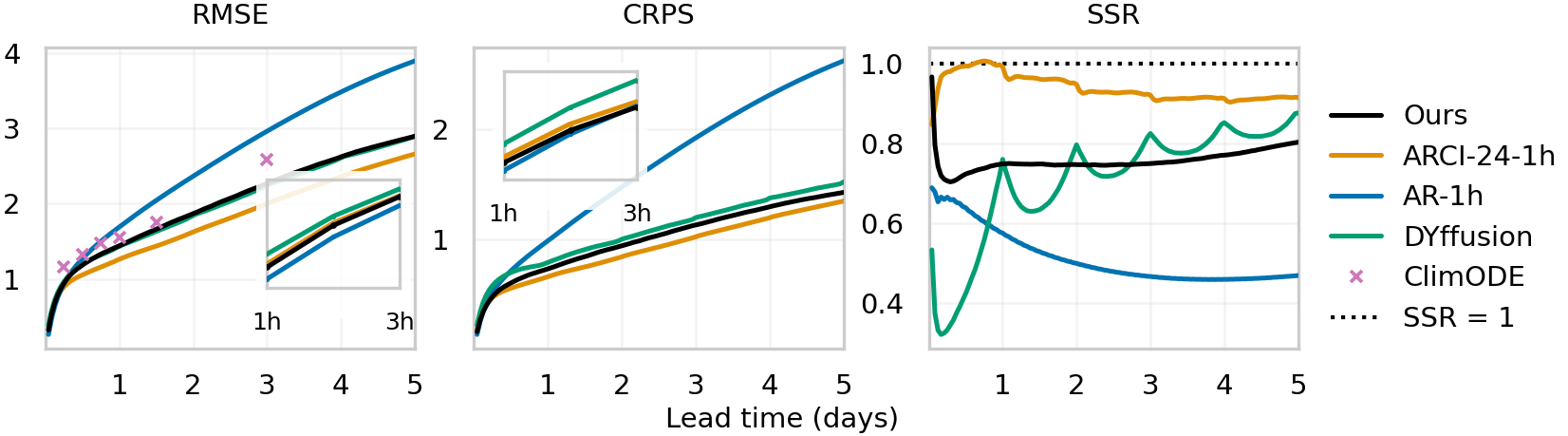}
  \caption{The RMSE ($\downarrow$), CRPS ($\downarrow$), SSR of \texttt{t850} for experiments on ERA5.}
  \label{fig:rmse_crps_ssr_era5}
\end{figure}

\begin{figure}[]
  \centering
  \setlength{\baselineskip}{0pt}
  \setlength{\lineskip}{0pt}
\includegraphics[width=\textwidth, trim={0 0 0 0}, clip]{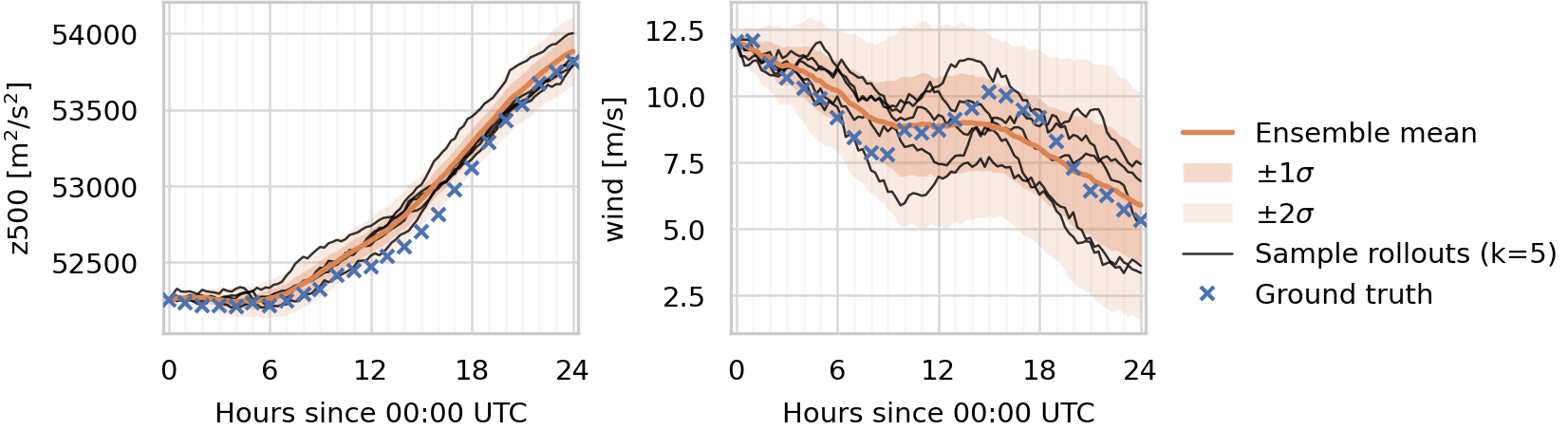}
  \caption{Continuity of dynamics in Northwest Europe ERA5 square. Mean and standard deviation (denoted $\sigma$) are computed from a 24-member rollout.}
  \label{fig:cont2}
\end{figure}

We additionally evaluate SDECast on global weather forecasting at one-hour resolution for lead times of up to five days. We use the ERA5 reanalysis dataset \cite{hersbach2020era5}, spatially downsampled to a resolution of $5.625^\circ$. We consider five atmospheric variables: 500\,hPa geopotential (\texttt{z500}), 850\,hPa temperature (\texttt{t850}), 2\,m temperature (\texttt{t2m}), and the 10\,m wind components (\texttt{u10} and \texttt{v10}).

We compare SDECast with four data-driven and continuous-time forecasting approaches. \textbf{AR-1h} is a conditional diffusion model trained to predict one-hour transitions and rolled out autoregressively. \textbf{ARCI-24-1h} \cite{AndraeLOL25} is trained directly across lead times from one to 24 hours. \textbf{DYffusion} \cite{cachay2023dyffusion} is trained on the same range of lead times and uses an iterative forecasting procedure. Finally, \textbf{ClimODE} \cite{verma2024climode} learns continuous-time weather dynamics using a physics-informed Neural ODE. All models are trained and evaluated using hourly ERA5 data, although we report ClimODE only from six-hour lead times because results at shorter lead times are unavailable.

We use the same SDECast parameterization as in the SQG experiment, but condition the drift and diffusion on the time of day and time of year to account for diurnal and seasonal variability. We further allow the diffusion coefficient $\vg_\theta$ to vary spatially, enabling the model to represent differences in forecast uncertainty across locations, such as between land and ocean. The learned diffusion magnitude is visualized in \cref{fig:vol_mag} and exhibits meaningful spatial variation.

As shown in \cref{fig:rmse_crps_ssr_era5}, SDECast produces stable and skillful probabilistic forecasts throughout the evaluation horizon.
Crucially, despite being trained on the same hourly data, SDECast substantially outperforms the one-hour autoregressive baseline at longer lead times.
This demonstrates that its continuous-time formulation supports high temporal resolution without suffering from the severe error accumulation observed in repeated one-hour autoregressive rollouts.
SDECast further improves on ClimODE and is comparable to DYffusion, while remaining behind the specialized ARCI-24-1h model.
This difference may be partly explained by the model inputs: SDECast and DYffusion assume Markovian dynamics and propagate a single state, whereas ARCI-24-1h conditions on the two previous states, providing access to additional temporal information.

To examine the learned continuous-time behavior more closely, we visualize SDECast trajectories for a single spatial location in \cref{fig:cont2}. The generated trajectories follow the overall evolution of the ground truth while exhibiting meaningful ensemble variation, illustrating how SDECast represents both temporal evolution and forecast uncertainty at an hourly resolution.

\section{Conclusions}

We introduced \textbf{SDECast}, a simulation-free Neural SDE framework for continuous-time probabilistic weather forecasting. 
By adapting SDE Matching to physical-space dynamics, SDECast learns explicit drift and diffusion functions from discretely observed trajectories and generates forecast ensembles at arbitrary temporal resolutions. 

Experiments on SQG and ERA5 demonstrate SDECast's ability to recover continuous-time dynamics and scale to global weather forecasting, highlighting Neural SDEs as a promising alternative to fixed-step autoregressive models. 
Current limitations include simplified posterior and diffusion-coefficient parameterizations and a limited experimental scope, motivating future work in this direction.

\begin{ack}
This research is financially supported by the Swedish Research Council (grant no: 2024-05011), the Wallenberg AI, Autonomous Systems and Software Program (WASP) funded by the Knut and Alice Wallenberg Foundation, and the Excellence Center at Linköping--Lund in Information Technology (ELLIIT). Our computations were enabled by the Berzelius resource at the National Supercomputer Centre, provided by the  Knut and Alice Wallenberg Foundation.
\end{ack}

\section*{References}
\printbibliography[heading=none]

\newpage
\appendix

\section{Details on SDE Matching}
\label{app:sde-matching}
For each $t$ and $\mat{X}$, the map $\vF_\varphi$ transports samples from $p(\vepsilon)$ to the posterior marginal $q_\varphi(\vz_t\mid\mat{X})$. Provided that $\vF_\varphi$ is differentiable with respect to $t$ and $\vepsilon$, and invertible with respect to $\vepsilon$, the posterior process admits an SDE representation whose drift is given by
\begin{align} 
\vf_{\varphi,\theta}(\vz_t,t;\mat{X}) =  \dot{\vF}_\varphi(\vF_\varphi^{-1}(\vz_t,t,\mat{X}),t,\mat{X}) + \vG_\theta(\vz_t,t) \nabla_{\vz_t} \log q_\varphi(\vz_t\mid\mat{X}) + \nabla_{\vz_t}\cdot \vG_\theta(\vz_t,t) 
\label{eq:posterior-drift} 
\end{align} 
where $\vG_\theta(\vz_t,t)=\frac{1}{2}\vg(\vz_t,t)\vg_\theta^\top(\vz_t,t)$.
Consequently, samples and posterior-drift evaluations can be obtained from $\vF_\varphi$ without numerically integrating \eqref{eq:posterior-sde} during training.
Since $\vF_\varphi$ is invertible with respect to $\vepsilon$, the score can be computed via the change-of-variables formula
\begin{align}
\nabla_{\vz_t}\log q_{\varphi}(\vz_t \mid \mat{X})
=   
\nabla_{\vz_t}\left[
\log p(\vepsilon)\Big|_{\vepsilon = \vF_{\varphi}^{-1}(\vz_t, t; \mat{X})}
+
\log\left|\det \frac{\partial\vF_{\varphi}^{-1}(\vz_t, t; \mat{X})}{\partial \vz_t}\right|
\right].
\label{eq:posterior_drift_expression}
\end{align}
All relevant quantities can be computed using automatic differentiation.
For our choice of $\vF_\varphi$ as
\begin{align} 
\vF_\varphi( \vepsilon,t, \mat{X}) 
= 
\bar{\vmu}_\varphi(t,\mat{X}) 
+ \bar{\vsigma}_\varphi(t,\mat{X}) \vepsilon,
\end{align}
the score admits the simple closed-form expression
\begin{align}
    \nabla_{\vz_t}\log q_{\varphi}(\vz_t\mid\mat{X}) = 
[\bar{\vsigma}_\varphi(\mat{X},t)\bar{\vsigma}_\varphi(\mat{X},t)^\top]^{-1}(\bar{\vmu}_\varphi(\mat{X},t) - \vz_t) = 
    -\bar{\vsigma}_\varphi(\mat{X},t)^{-1}\vepsilon.
    \label{eq:score}
\end{align}
Finally, whenever $\vg_\theta$ is state-independent, the divergence term $\nabla_{\vz_t}\cdot \vG_\theta(\vz_t,t)$ in \cref{eq:posterior-drift} vanishes.

\subsection{SDECast algorithm}
\label{sec:algorithm}

\Cref{alg:sdecast} summarises one training step of SDECast with the \textbf{learnable}
posterior of \cref{eq:local-posterior-map}. On each interval $[t_i,t_{i+1}]$ the posterior marginal
$q_{\varphi}(\bm{z}_t \mid \bm{x}_{t_i}, \bm{x}_{t_{i+1}})$ is Gaussian: its mean is the linear
interpolation of the two bracketing observations plus a learnable correction
$\hl{orange}{\bm{\mu}_{\varphi}}$, and its scale is $\delta$ plus a learnable correction
$\hl{pink}{\bm{\sigma}_{\varphi}}$. Both corrections are multiplied by
$\hl{yellow}{\alpha_t(1-\alpha_t)}$ and therefore vanish at the observation times, anchoring
$\bm{z}_{t_i}\sim\mathcal{N}(\bm{x}_{t_i},\delta^2\bm{I})$ as required by \cref{eq:observation-model}.
The simplified \textbf{linear-interpolant} variant is recovered by fixing
$\hl{orange}{\bm{\mu}_{\varphi}=\bm{0}}$ and $\hl{pink}{\bm{\sigma}_{\varphi}=\bm{1}}$, which
leaves a Brownian-bridge-shaped posterior with no learnable parameters.

\begin{algorithm}[H]
\caption{SDECast training step}
\label{alg:sdecast}
\begin{algorithmic}[1]
\REQUIRE $\mathbf{X}=\{\bm{x}_{t_i}\}_{i=0}^{N}$ observed at $0=t_0<\dots<t_N=T$
\REQUIRE $\bm{h}_{\theta},\bm{g}_{\theta}$ -- prior drift and diffusion networks
\REQUIRE $\bm{\mu}_{\varphi},\bm{\sigma}_{\varphi}$ -- learnable posterior corrections
\WHILE{not converged}
    \STATE $i \sim \mathcal{U}\{0,\dots,N-1\}$, \quad
           $\alpha \sim \mathcal{U}(0,1)$, \quad
           $\bm{\varepsilon} \sim \mathcal{N}(\bm{0},\bm{I})$
    \STATE $t \gets t_i + \alpha\,(t_{i+1}-t_i)$
      \COMMENT{physical time}
    \STATE \mycomment{Sample the posterior marginal $\bm{z}_t=\bm{F}_{\varphi}(\bm{\varepsilon},t,\mathbf{X})$ of \cref{eq:local-posterior-map}}
    \STATE $\bar{\bm{\mu}}_{\varphi} \gets (1-\alpha_t)\bm{x}_{t_i} + \alpha_t\bm{x}_{t_{i+1}}
            + \hl{yellow}{\alpha_t(1-\alpha_t)}\,\hl{orange}{\bm{\mu}_{\varphi}(t,\bm{x}_{t_i},\bm{x}_{t_{i+1}})}$
    \STATE $\bar{\bm{\sigma}}_{\varphi} \gets \delta
            + \hl{yellow}{\alpha_t(1-\alpha_t)}\,\hl{pink}{\bm{\sigma}_{\varphi}(t,\bm{x}_{t_i},\bm{x}_{t_{i+1}})}$
    \STATE $\bm{z}_t \gets \bar{\bm{\mu}}_{\varphi} + \bar{\bm{\sigma}}_{\varphi}\bm{\varepsilon}$
    \STATE \mycomment{Posterior drift at $\bm{z}_t$ via \cref{eq:posterior-drift}}
    \STATE $\dot{\bm{F}}_{\varphi} \gets \partial \bm{z}_t / \partial t$
      \COMMENT{autograd}
    \STATE $\bm{G}_{\theta} \gets \tfrac{1}{2}\bm{g}_{\theta}(\bm{z}_t,t)\bm{g}_{\theta}^{\top}(\bm{z}_t,t)$
    \STATE $\nabla_{\bm{z}_t}\log q_{\varphi}(\bm{z}_t\mid\mathbf{X})
            \gets -\bar{\bm{\sigma}}_{\varphi}^{-1}\bm{\varepsilon}$
      \COMMENT{\cref{eq:score}}
    \STATE $\bm{f}_{\varphi,\theta} \gets \dot{\bm{F}}_{\varphi}
            + \bm{G}_{\theta}\nabla_{\bm{z}_t}\log q_{\varphi}(\bm{z}_t\mid\mathbf{X})
            + \nabla_{\bm{z}_t}\!\cdot\bm{G}_{\theta}$
    \STATE \mycomment{Control and simulation-free dynamics loss}
    \STATE $\bm{u}_{\theta,\varphi}(\bm{z}_t,t,\mathbf{X}) \gets
            \bm{g}_{\theta}^{-1}(\bm{z}_t,t)\bigl(\bm{f}_{\varphi,\theta}(\bm{z}_t,t;\mathbf{X})
            - \bm{h}_{\theta}(\bm{z}_t,t)\bigr)$
    \STATE $\hat{\mathcal{L}}_{\mathrm{dyn}} \gets
            \tfrac{1}{2}\bigl\|\bm{u}_{\theta,\varphi}(\bm{z}_t,t,\mathbf{X})\bigr\|_2^2$
    \STATE Gradient step on $\theta,\varphi$ w.r.t.\ $\hat{\mathcal{L}}_{\mathrm{dyn}}$
\ENDWHILE
\end{algorithmic}
\end{algorithm}

\section{Experimental Details}
\label{app:experiments}

\subsection{ERA5}

We train SDECast on the WeatherBench-1 \cite{rasp2023weatherbench2} copy of
ERA5 at $5.625^\circ$ resolution, a $32\times64$ latitude--longitude grid, on the
five variables $z500$, $t850$, $t2m$, $u10$ and $v10$, each standardised to zero
mean and unit variance using statistics computed over the training period
(1979--2015; validation 2016, test 2018). Longitude is periodic, so the
convolutions wrap along the longitude axis only.

The drift and interpolant networks receive $12$ conditioning channels
($6$ spatial, $2$ static surface, $4$ temporal). The
input to the drift network is therefore the $5$ state channels plus $12$
conditioning channels; the interpolant network takes the two endpoint states
($2\times5$) plus the same $12$ conditioning channels. The diffusion field
$g_\theta$ has $5\times32\times64$ entries. Parameter counts are given
in Table~\ref{tab:era5}. During inference, the SDE is solved with 64 Euler-Maruyama steps.

For details on the baseline implementations we refer to \cite{AndraeLOL25}.

\begin{table}[b]
  \centering
  \caption{ERA5 model configuration ($5.625^\circ$, $32\times64$ grid).}
  \label{tab:era5}
  \begin{tabular}{@{}ll@{}}
    \toprule
    Component & Value \\
    \midrule
    Grid (lat $\times$ lon)            & $32 \times 64$ \\
    State variables $C$                & $5$ ($z500,t850,t2m,u10,v10$) \\
    Conditioning channels              & $12$ ($6$ spatial $+ 2$ static $+ 4$ temporal) \\
    Padding                            & circular in longitude only \\
    Posterior interpolant              & learnable \\
    Volatility $g_\theta$              & state-independent, scalar per variable and grid cell \\
    \midrule
    Drift network $h_\theta$           & $3{,}556{,}741$ params \\
    Interpolant network $F_\varphi$    & $3{,}563{,}946$ params \\
    Diffusion field $g_\theta$         & $10{,}240$ params ($5\times32\times64$) \\
    \bottomrule
  \end{tabular}
\end{table}

\subsection{SQG}
\label{app:arch:sqg}

For the SQG experiments the state is the scaled surface potential temperature on
the two bounding surfaces ($z=0$ and $z=H$), giving $C=2$ channels on a
$64\times64$ doubly-periodic grid; the convolutions therefore wrap along both
spatial axes. There is no extra static or temporal conditioning. We use the same backbone and model size
as for ERA5 (Table~\ref{tab:era5}) for prior drift and learnable interpolant, but parametrize the diffusion coefficient as a single scalar. We train models with both the linear and the learnable interpolant. 
During inference, the SDE is solved with 200 Euler-Maruyama steps.

\begin{table}
  \centering
  \caption{SQG model configuration ($64\times64$ doubly-periodic grid).}
  \label{tab:sqg}
  \begin{tabular}{@{}ll@{}}
    \toprule
    Component & Value \\
    \midrule
    Grid                               & $64 \times 64$ (doubly periodic) \\
    State channels $C$                 & $2$ (surfaces $z=0,\,z=H$) \\
    Conditioning                       & none \\
    Padding                            & circular in both axes \\
    Posterior interpolant              & linear / learnable \\
    Volatility $g_\theta$              & state-independent, one scalar \\
    \midrule
    Drift network $h_\theta$           & $3{,}542{,}050$ params \\
    Interpolant network $F_\varphi$    & $3{,}544{,}932$ params \\
    Diffusion field $g_\theta$         & $8{,}192$ params ($2\times64\times64$) \\
    \bottomrule
  \end{tabular}
\end{table}

\section{Training and Optimisation}
\label{app:arch:train}

SDECast is trained with the  simulation-free objective
in \cref{eq:loss} using AdamW, with a linear
learning-rate warmup followed by cosine annealing.
The $\delta$ parameter is chosen to be small enough to not affect the results, while still being non-zero to avoid division by zero in \cref{eq:score}. This equates to $\delta=0.001$ in both experiments.
Optimisation settings are summarised in Table~\ref{tab:train}.

\begin{table}[t]
  \centering
  \caption{Training and optimisation settings.}
  \label{tab:train}
  \begin{tabular}{@{}lll@{}}
    \toprule
    Setting & ERA5 & SQG \\
    \midrule
    Optimiser                       & AdamW & AdamW \\
    Learning rate                   & $5\times10^{-4}$ & $5\times10^{-4}$ \\
    Weight decay                    & $1\times10^{-4}$ & $1\times10^{-4}$ \\
    LR warmup                       & $500$ steps, linear & $500$ steps, linear \\
    LR schedule                     & cosine $\rightarrow 10^{-5}$ & cosine $\rightarrow 10^{-5}$ \\
    Epochs                          & $50$ & $20$  \\
    Batch size                      & $128$  & $128$  \\
    Precision                       & fp32 & fp32 \\
    Hardware                        & $2\times$ A100 ($\sim$24\,h) & $2\times$ H100 ($\sim$7\,h)  \\
    \midrule
    Forecast lead time              & $1$\,h & $6$\,h  \\
    Loss weighting                  & uniform & uniform \\
    \bottomrule
  \end{tabular}
\end{table}

\section{Additional results on ERA5}
\begin{figure}[]
  \centering
  \setlength{\baselineskip}{0pt}
  \setlength{\lineskip}{0pt}
\includegraphics[width=0.49\textwidth, trim={0 0 0 0}, clip]{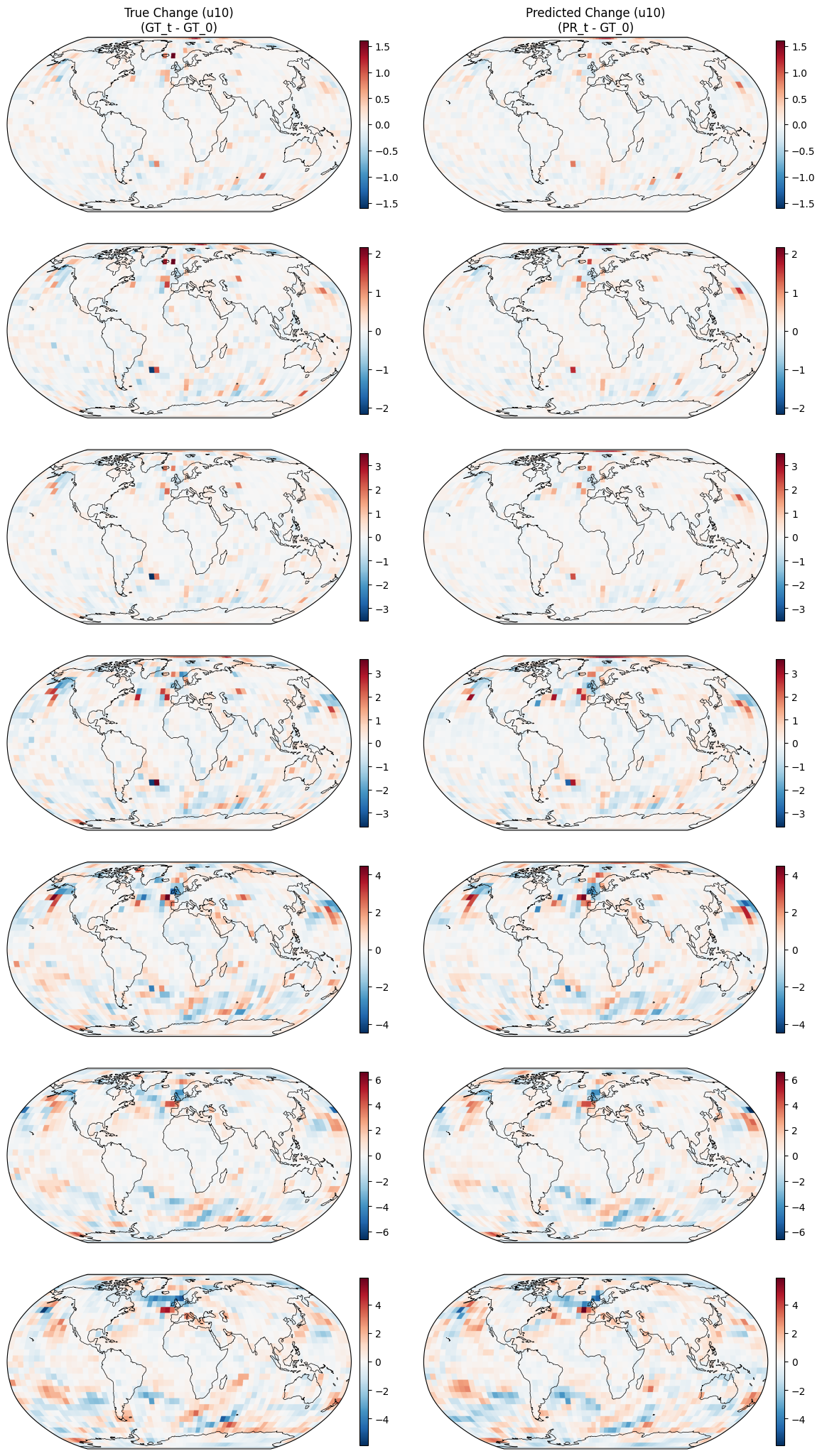}
\includegraphics[width=0.49\textwidth, trim={0 0 0 0}, clip]{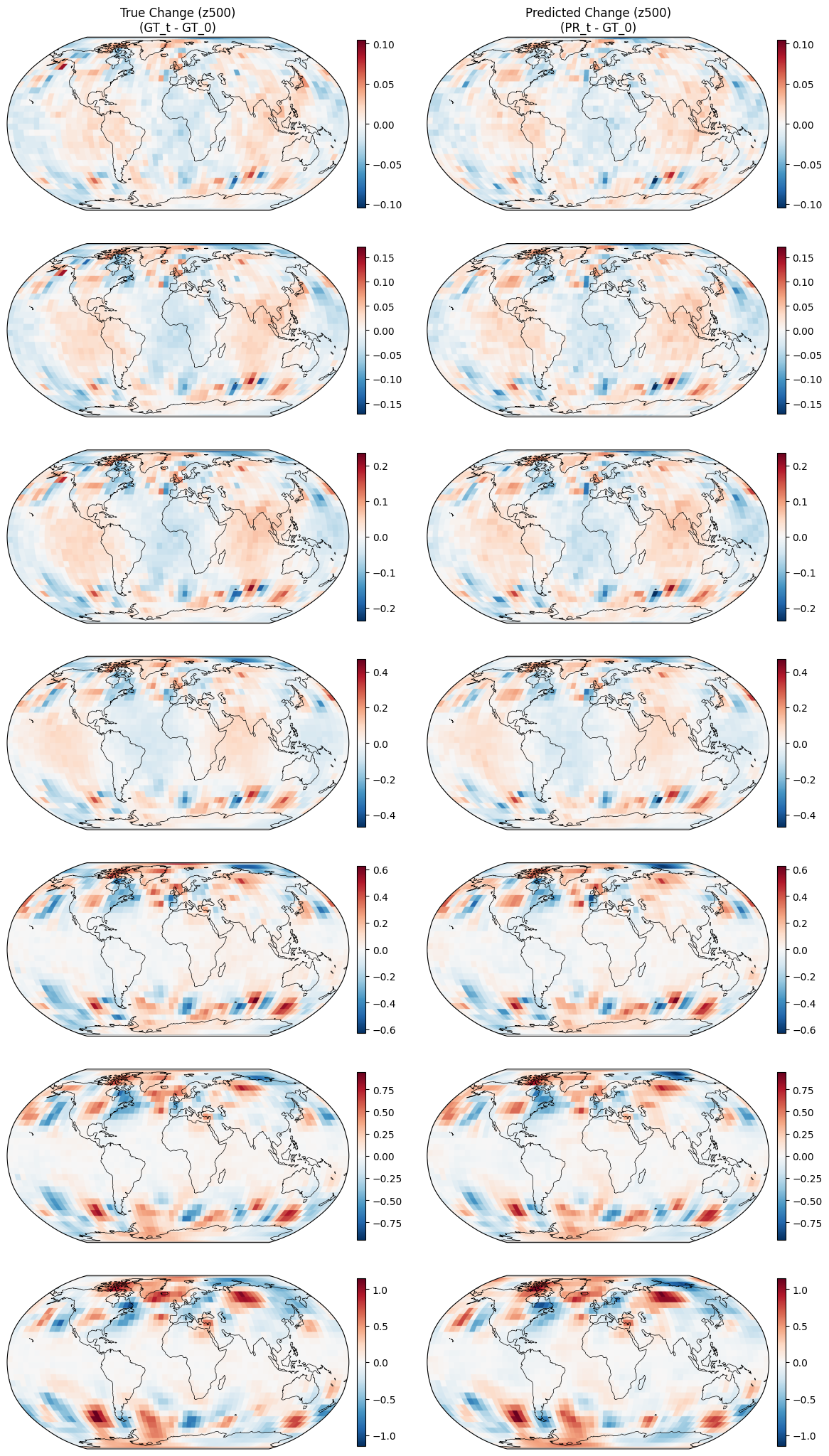}
  \caption{Predicted and ground truth change in u10 and z500 for the following lead times (from top to bottom): 1, 2, 3, 6, 12, 24 and 48 hours. Predicted changes were obtained from a single trajectory.}
  \label{fig:residuals}
\end{figure}

\begin{figure}[]
  \centering
  \newlength{\imghtx}
\settoheight{\imghtx}{\includegraphics{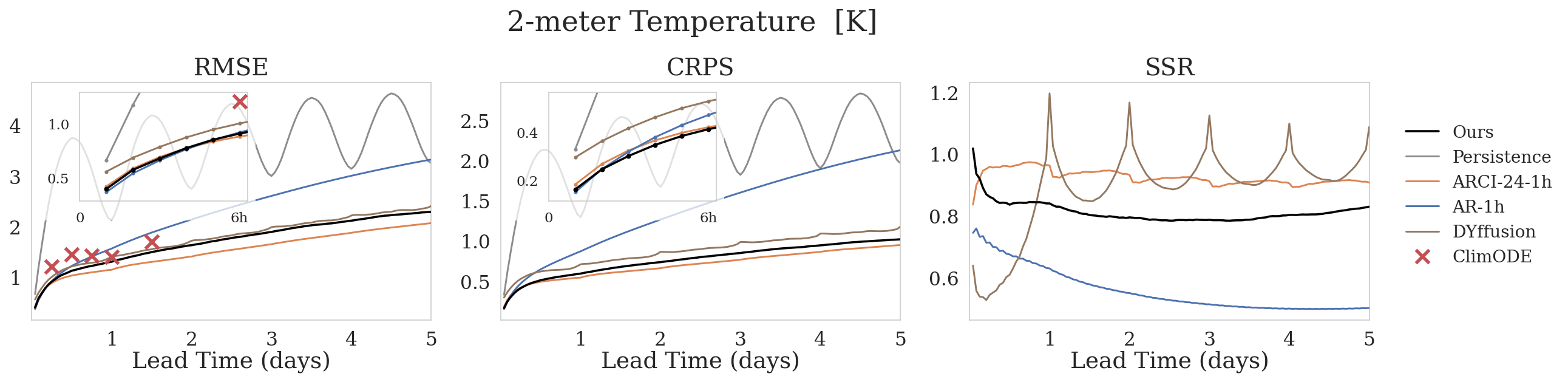}}
  \setlength{\baselineskip}{0pt}
  \setlength{\lineskip}{0pt}
\includegraphics[width=0.9\textwidth, trim={0 0 0 0}, clip]{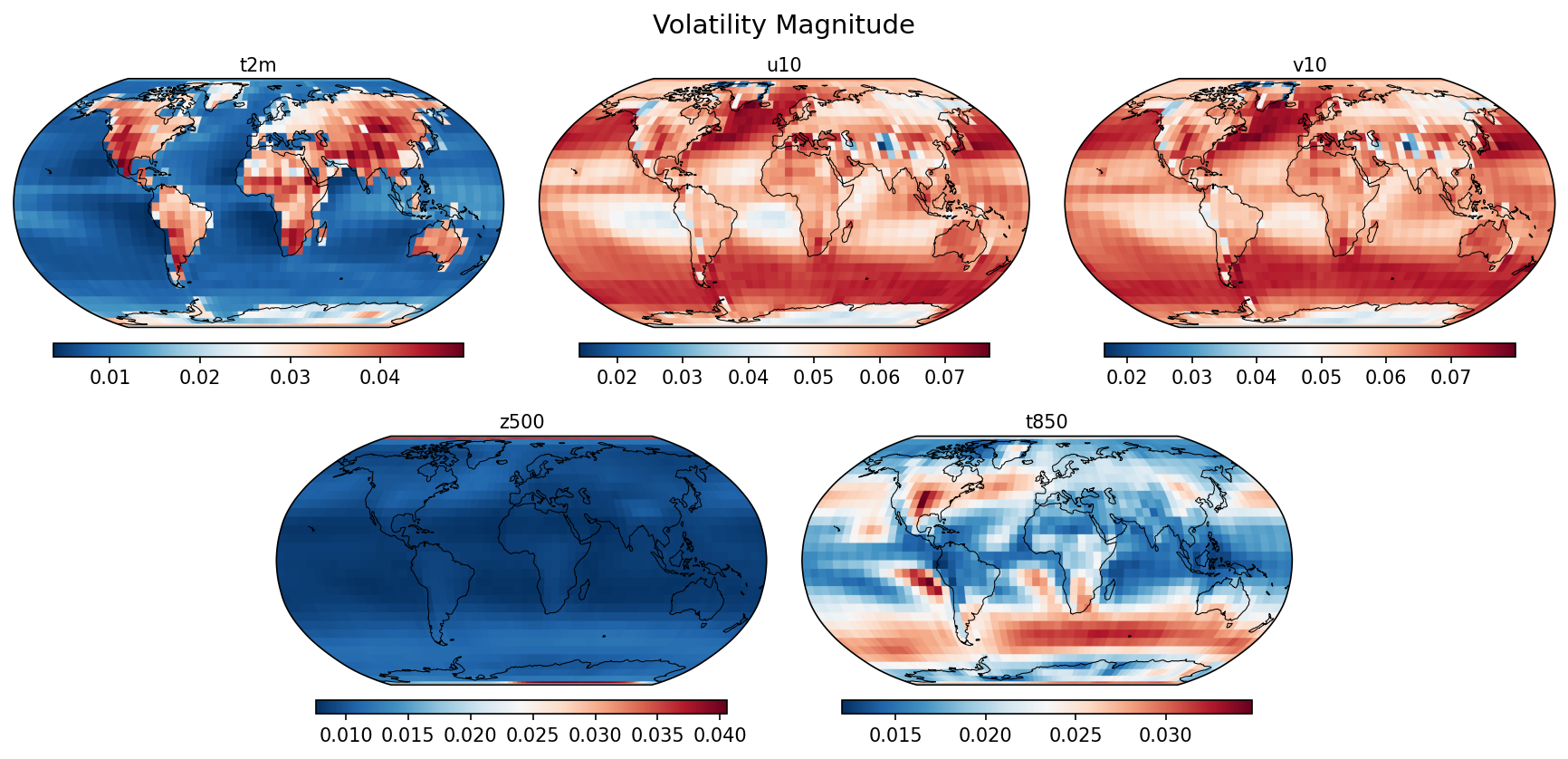}
  \caption{Learned volatility magnitude visualization.}
  \label{fig:vol_mag}
\end{figure}

\begin{figure}[]
  \centering
  \setlength{\baselineskip}{0pt}
  \setlength{\lineskip}{0pt}
\includegraphics[width=\textwidth, trim={0 0 0 0}, clip]{workshop_figs/w_era5_metrics_zoom1-3l_t850.png}
\includegraphics[width=\textwidth, trim={0 0 0 0}, clip]{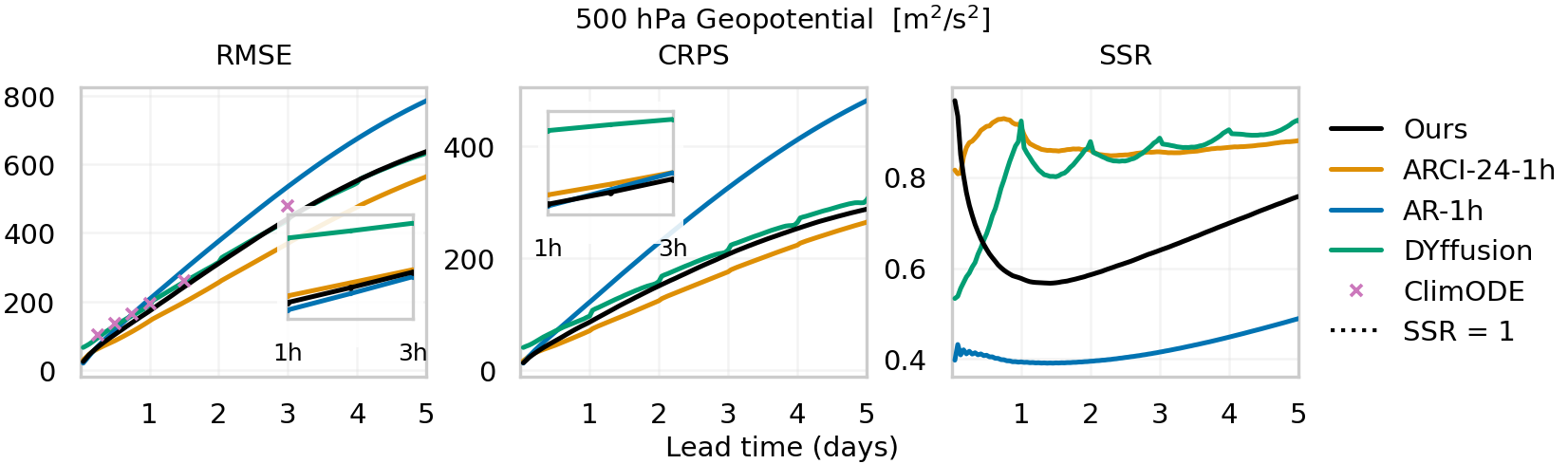}
\includegraphics[width=\textwidth, trim={0 0 0 0}, clip]{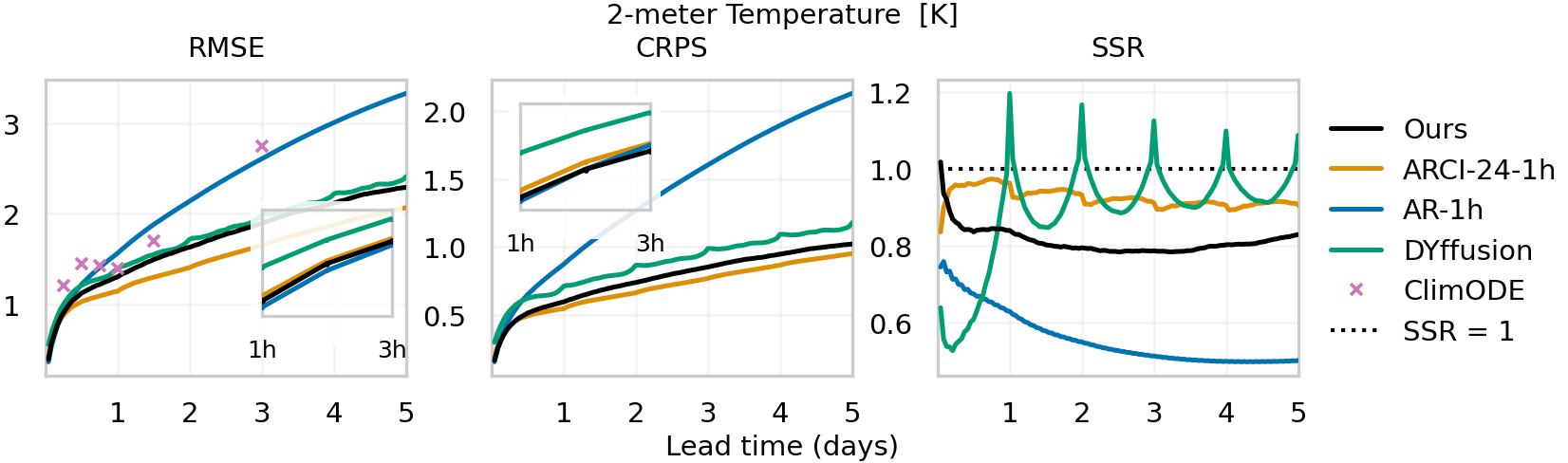}
\includegraphics[width=\textwidth, trim={0 0 0 0}, clip]{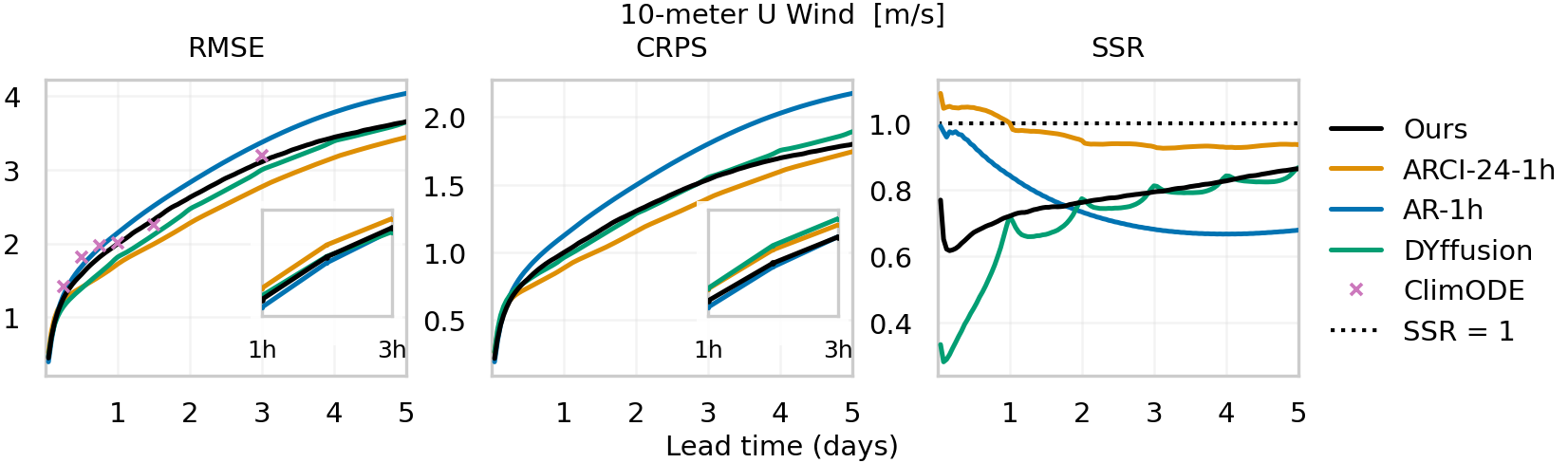}
\includegraphics[width=\textwidth, trim={0 0 0 0}, clip]{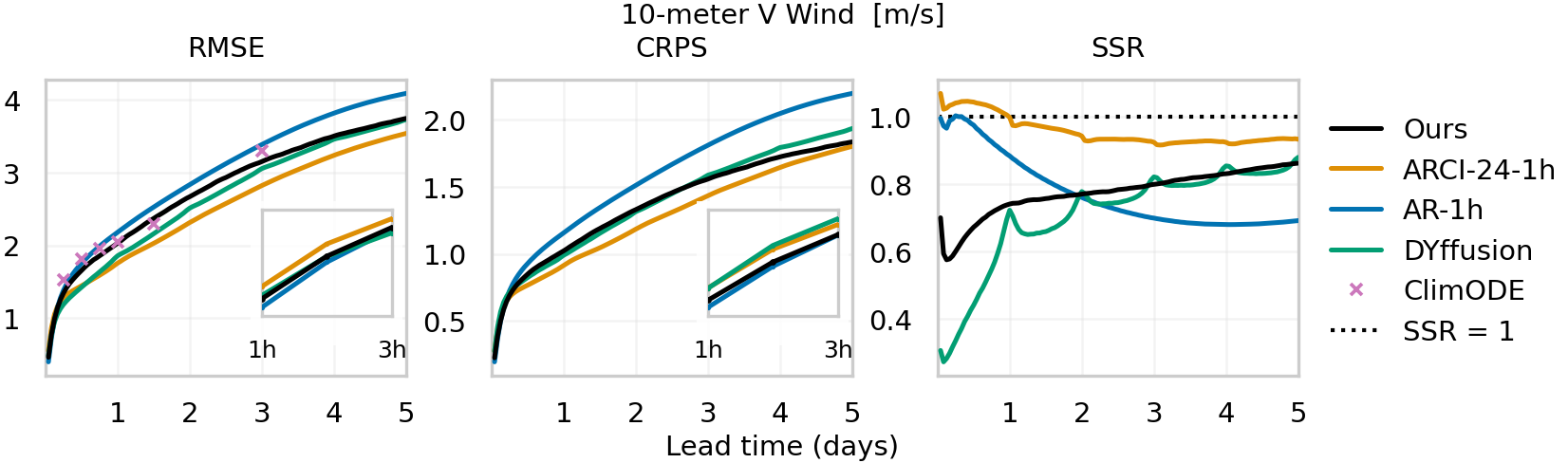}
  \caption{Latitude-weighted RMSE, CRPS and SSR. RMSE and CRPS are given in physical units; z500 in m$^2$/s$^2$, t850/t2m in K, u10/v10 in m/s) at 1, 6 and 24\,h lead times. ClimODE paper does not report its CRPS and SSR so we omit it from the corresponding plots.}
  \label{fig:rmse_crps_ssr_era5_large}
\end{figure}

We provide additional visualizations for the ERA5 experiments including forecast residuals in \cref{fig:residuals} and learned volatility in \cref{fig:vol_mag}. Additional scores for all variables are also provided in \cref{tab:rmse-era5,tab:crps-era5} and \cref{fig:rmse_crps_ssr_era5_large}.

\begin{table}
\centering
\caption{Latitude-weighted RMSE ($\downarrow$) on ERA5 in physical units (z500 in m$^2$/s$^2$, t850/t2m in K, u10/v10 in m/s) at 1, 6 and 24\,h lead times.}
\label{tab:rmse-era5}
\begin{tabular}{llcccccc}
\toprule
Variable & Time (h) & SDECast & Persistence & ARCI-24-1h & AR-1h & DYffusion & ClimODE \\
\midrule
\multirow{3}{*}{z500} & 1 & \underline{24.8} & 50.9 & 29.0 & \textbf{20.0} & 65.8 & -- \\
 & 6 & \underline{66.7} & 235.3 & \textbf{61.7} & 68.5 & 93.9 & 102.9 \\
 & 24 & \underline{172.0} & 603.1 & \textbf{142.9} & 206.7 & 186.5 & 193.4 \\
\midrule
\multirow{3}{*}{t850} & 1 & \underline{0.32} & 0.40 & 0.34 & \textbf{0.26} & 0.39 & -- \\
 & 6 & 0.94 & 1.59 & \textbf{0.89} & \underline{0.93} & 0.96 & 1.16 \\
 & 24 & 1.44 & 3.09 & \textbf{1.26} & 1.69 & \underline{1.43} & 1.55 \\
\midrule
\multirow{3}{*}{t2m} & 1 & \underline{0.40} & 0.66 & 0.42 & \textbf{0.37} & 0.56 & -- \\
 & 6 & \underline{0.91} & 2.76 & \textbf{0.88} & 0.92 & 1.00 & 1.21 \\
 & 24 & \underline{1.30} & 2.13 & \textbf{1.14} & 1.57 & 1.40 & 1.40 \\
\midrule
\multirow{3}{*}{u10} & 1 & \underline{0.44} & 0.59 & 0.52 & \textbf{0.39} & 0.47 & -- \\
 & 6 & 1.26 & 2.22 & \underline{1.22} & 1.30 & \textbf{1.14} & 1.41 \\
 & 24 & 1.99 & 4.06 & \textbf{1.72} & 2.14 & \underline{1.81} & 2.01 \\
\midrule
\multirow{3}{*}{v10} & 1 & \underline{0.46} & 0.65 & 0.55 & \textbf{0.40} & 0.49 & -- \\
 & 6 & 1.34 & 2.51 & \underline{1.28} & 1.38 & \textbf{1.19} & 1.53 \\
 & 24 & 2.03 & 4.71 & \textbf{1.75} & 2.18 & \underline{1.86} & 2.04 \\
\bottomrule
\end{tabular}
\end{table}

\begin{table}
\centering
\caption{Latitude-weighted CRPS ($\downarrow$) on ERA5 in physical units (z500 in m$^2$/s$^2$, t850/t2m in K, u10/v10 in m/s) at 1, 6 and 24\,h lead times. ClimODE is deterministic and is therefore not evaluated here.}
\label{tab:crps-era5}
\begin{tabular}{llccccc}
\toprule
Variable & Time (h) & SDECast & Persistence & ARCI-24-1h & AR-1h & DYffusion \\
\midrule
\multirow{3}{*}{z500} & 1 & \underline{13.4} & 38.6 & 16.8 & \textbf{12.7} & 40.7 \\
 & 6 & \underline{34.0} & 169.2 & \textbf{33.1} & 40.6 & 53.5 \\
 & 24 & \underline{86.1} & 352.6 & \textbf{70.6} & 121.7 & 96.7 \\
\midrule
\multirow{3}{*}{t850} & 1 & \underline{0.16} & 0.26 & 0.18 & \textbf{0.13} & 0.22 \\
 & 6 & \underline{0.48} & 1.06 & \textbf{0.47} & 0.52 & 0.58 \\
 & 24 & \underline{0.73} & 2.02 & \textbf{0.66} & 0.98 & 0.79 \\
\midrule
\multirow{3}{*}{t2m} & 1 & \underline{0.16} & 0.33 & 0.18 & \textbf{0.15} & 0.30 \\
 & 6 & \textbf{0.41} & 1.51 & \underline{0.42} & 0.47 & 0.53 \\
 & 24 & \underline{0.59} & 1.30 & \textbf{0.54} & 0.87 & 0.71 \\
\midrule
\multirow{3}{*}{u10} & 1 & \underline{0.22} & 0.37 & 0.26 & \textbf{0.19} & 0.26 \\
 & 6 & \textbf{0.64} & 1.45 & \underline{0.64} & 0.68 & 0.68 \\
 & 24 & 1.00 & 2.64 & \textbf{0.88} & 1.12 & \underline{0.96} \\
\midrule
\multirow{3}{*}{v10} & 1 & \underline{0.22} & 0.39 & 0.28 & \textbf{0.19} & 0.27 \\
 & 6 & \underline{0.69} & 1.61 & \textbf{0.68} & 0.73 & 0.71 \\
 & 24 & 1.02 & 3.07 & \textbf{0.91} & 1.16 & \underline{0.99} \\
\bottomrule
\end{tabular}
\end{table}

\clearpage
\section{Additional results on SQG}
Additional scores for SQG are provided in \cref{tab:rmse,tab:crps,tab:ssr}.

\begin{table}[H]
  \centering
  \small
  \caption{Root-mean-square error (RMSE) by forecast lead time, on SQG in physical
  (denormalized) units. Values are the mean over 24 test trajectories.}
  \label{tab:rmse}
  \begin{tabular*}{\textwidth}{@{\extracolsep{\fill}}>{\raggedright\arraybackslash}p{4cm}cccccccc@{}}
    \toprule
    & & \multicolumn{7}{c}{RMSE\,$\downarrow$} \\
    \cmidrule(lr){3-9}
    Model & Lead time & 1\,h & 3\,h & 6\,h & 12\,h & 24\,h & 48\,h & 96\,h \\
    \midrule
    Persistence & -- & 1.635 & 3.803 & 5.319 & 6.756 & 7.967 & 8.925 & 9.665 \\
    \midrule
    CFM & 6\,h & -- & -- & \textbf{0.832} & \textbf{1.107} & \textbf{1.664} & \textbf{3.004} & \textbf{5.301} \\
    \midrule
    SDECast (linear) & 6\,h & \underline{1.494} & \underline{2.733} & 2.284 & 2.873 & 3.918 & 5.535 & 7.726 \\
    \midrule
    SDECast & 6\,h & \textbf{0.561} & \textbf{1.055} & \underline{1.280} & \underline{1.584} & \underline{2.277} & \underline{3.803} & \underline{5.863} \\
    \bottomrule
  \end{tabular*}
\end{table}

\begin{table}[H]
  \centering
  \small
  \caption{Continuous ranked probability score (CRPS) by forecast lead time, on SQG in
  physical (denormalized) units. Values are the mean over 24 test trajectories.}
  \label{tab:crps}
  \begin{tabular*}{\textwidth}{@{\extracolsep{\fill}}llccccccc@{}}
    \toprule
    & & \multicolumn{7}{c}{CRPS $\downarrow$} \\
    \cmidrule(lr){3-9}
    Model & Lead time & 1\,h & 3\,h & 6\,h & 12\,h & 24\,h & 48\,h & 96\,h \\
    \midrule
    Persistence & -- & 1.150 & 2.741 & 3.915 & 5.107 & 6.188 & 7.077 & 7.771 \\
    \midrule
    CFM & 6\,h & -- & -- & \textbf{0.441} & \textbf{0.591} & \textbf{0.879} & \textbf{1.561} & \textbf{2.822} \\
    \midrule
    SDECast (linear) & 6\,h & \underline{0.907} & \underline{1.729} & 1.402 & 1.677 & 2.172 & 2.989 & 4.112 \\
    \midrule
    SDECast & 6\,h & \textbf{0.303} & \textbf{0.599} & \underline{0.717} & \underline{0.868} & \underline{1.195} & \underline{1.956} & \underline{3.163} \\
    \bottomrule
  \end{tabular*}
\end{table}

\begin{table}[H]
  \centering
  \small
  \caption{Spread--skill ratio (SSR) by forecast lead time (dimensionless;
  $\mathrm{SSR}=1$ indicates a calibrated ensemble). Values are the mean over 24 test trajectories. This is an ensemble-only metric, so the
  deterministic baseline (Persistence) report $0$ and are excluded from
  highlighting; dashes denote leads the model does not produce.}
  \label{tab:ssr}
  \begin{tabular*}{\textwidth}{@{\extracolsep{\fill}}>{\raggedright\arraybackslash}p{4cm}cccccccc@{}}
    \toprule
    & & \multicolumn{7}{c}{SSR} \\
    \cmidrule(lr){3-9}
    Model & Lead time & 1\,h & 3\,h & 6\,h & 12\,h & 24\,h & 48\,h & 96\,h \\
    \midrule
    Persistence & -- & 0.000 & 0.000 & 0.000 & 0.000 & 0.000 & 0.000 & 0.000 \\
    \midrule
    CFM & 6\,h & -- & -- & \textbf{0.854} & \textbf{0.833} & \underline{0.810} & \underline{0.833} & \underline{0.935} \\
    \midrule
    SDECast (linear) & 6\,h & 0.211 & 0.179 & 0.334 & 0.440 & 0.583 & 0.817 & 1.419 \\
    \midrule
    SDECast & 6\,h & \textbf{0.622} & \textbf{0.552} & \underline{0.641} & 0.789 & \textbf{0.945} & \textbf{0.981} & \textbf{0.966} \\
    \bottomrule
  \end{tabular*}
\end{table}

\end{document}